\documentclass[10pt,conf]{IEEEtran}
\usepackage{geometry}
 \usepackage{bbold}
\usepackage{ifpdf}
\usepackage{subcaption}%
\usepackage[cmex10]{amsmath}
\usepackage{amssymb}
\usepackage{verbatim}
\usepackage{algorithm}
\usepackage{algorithmic}
\usepackage{array}
\usepackage{latexsym}
\usepackage{color}
\usepackage{textgreek}
\usepackage{subscript}
\usepackage{rotating}
\usepackage{booktabs,multirow}
\usepackage{mdframed}	
\usepackage{tabularx}
\usepackage{amsmath}
\usepackage{makecell}
\usepackage{tabto}
\usepackage{mathrsfs}
\usepackage{url}
\usepackage{cite}
\usepackage{listings}
\usepackage{array}
\usepackage[table]{xcolor}

\usepackage{microtype}

\newcommand{\nref}[1]{\textcolor{red}{[{\bf NEED CITATION}]}}
\begin{document}
\setlength{\columnsep}{0.2 in}
\def\BibTeX{{\rm B\kern-.05em{\sc i\kern-.025em b}\kern-.08em T\kern-.1667em\lower.7ex\hbox{E}\kern-.125emX}}

\title{Federated Learning-Based Risk-Aware Decision to Mitigate Fake Task Impacts on Crowdsensing  Platforms}
\author{
	Zhiyan~Chen,~Murat~Simsek,~\IEEEmembership{Senior~Member,~IEEE},~and~Burak Kantarci,~\IEEEmembership{Senior Member,~IEEE}
	\thanks{
	The authors are with the School of Electrical Engineering and 
        Computer Science at the University of Ottawa, Ottawa, ON, K1N 6N5, Canada.
		E-mail: \{zchen241, murat.simsek,burak.kantarci\}@uottawa.ca}  

\vspace{-0.3in}

}

\maketitle
\thispagestyle{empty}
\pagestyle{empty}

\begin{abstract}
\setlength{\columnsep}{0.2 in}
Mobile crowdsensing (MCS) leverages distributed and non-dedicated sensing concepts by utilizing sensors imbedded in a large number of mobile smart devices. However, the openness and distributed nature of MCS leads to various vulnerabilities and consequent challenges to address. A malicious user submitting fake sensing tasks to an MCS platform may be attempting to consume resources from any number of participants’ devices; as well as attempting to clog the MCS server. In this paper, a novel approach that is based on horizontal federated learning is proposed to identify fake tasks that contain a number of independent detection devices and an aggregation entity. Detection devices are deployed to operate in parallel with each device equipped with a machine learning (ML) module, and an associated training dataset. Furthermore, the aggregation module collects the prediction results from individual devices and determines the final decision with the objective of minimizing the prediction loss. Loss measurement considers the lost task values with respect to misclassification, where the final decision utilizes a risk-aware approach where the risk is formulated as a function of the utility loss. Experimental results demonstrate that using federated learning-driven illegitimate task detection with a risk aware aggregation function improves the detection performance of the traditional centralized framework. Furthermore, the higher performance of detection and lower loss of utility can be achieved by the proposed framework. This scheme can even achieve 100\% detection accuracy using small training datasets distributed across devices, while achieving slightly over an 8\% increase in detection improvement over traditional approaches.
\end{abstract}
\begin{IEEEkeywords}
\setlength{\columnsep}{0.2 in}
	Mobile Crowdsensing, Internet of Things, Machine Learning, risk-awareness, utility loss, Federated Learning.\end{IEEEkeywords}

\IEEEpeerreviewmaketitle

\section{Introduction}
\label{sec:intro}
Mobile crowdsensing (MCS) is a cloud-inspired model for sensing, clustering and aggregating data via smart devices (e.g., smart phones, tablets, and in-vehicle sensors) and becomes an engaging topic \cite{8703108}. 
Although MCS is applied in many areas, it confronts a number of security challenges and threats (e.g., data poisoning threat, privacy leakage and malware attack)\cite{6069707}. Among them, a fake task attack is one of the top crucial threats where adversaries aim to clog the MCS servers and also drain resources from the devices that participate in the MCS campaigns \cite{8969416}. Energy-oriented illegitimate tasks result in consuming excessive resources from users' equipment such as energy, bandwidth and computation capability, which are all limited in capacity in such smart devices\cite{Zhang.2020.iot}.
Furthermore, clogging MCS server via illegitimate task injection diminishes the effectiveness of the platform and suppresses users’ willingness to take part in MCS activities \cite{Zhang.2020.iot}. In order to protect both MCS platform and users from malicious activities of fake tasks, the studies in \cite{audrey.sose2019, Audrey.ACM.2019, simsek.2020}  proposed machine learning (ML)-based approaches to mitigate  fake tasks in MCS platforms.

An alternative to a centralized approach to detect the fake submissions is to decentralize the decision and offload detection to the participating devices, which yields Federated Learning (FL). This approach also mitigates the concerns about privacy in data provisioning to the centralized servers' machine learning models.
FL is a methodology integrating a distributed ML paradigm that trains a shared and dispensed artificial intelligence (AI) model using local training datasets on distributed devices \cite{xu2019verifynet}. Research on the application of FL to fake task detection is limited. In this paper, we propose a novel horizontal FL-backed framework to identify malicious tasks in MCS systems. The proposed system is comprised of various detection devices and a risk/loss-aware\footnote{We use risk-aware and risk/loss-aware interchangeably as risk is the cost of utility losses due to misclassifying a task under the wrong legitimacy class and losing a legitimate task value or incurring the cost of an illegitimate task.} aggregation module. A detection device runs a local ML model that may vary from one device to another (e.g., Bagging, Adaboost and Extreme Gradient Boosting (XGBoost)) and a local dataset. The machine learning model is trained by the local dataset in the device and delivers prediction results independently. The aggregation module is responsible for gathering decisions from devices and making a concluding decision. Bayesian decision theory suggests making statistic-driven decisions with the objective of maximizing the expected utility over a model posterior \cite{kusmierczyk2019variational}. With this in mind, we design an aggregation module that makes decisions subject to the minimum utility loss over the actions regarding the classes of incoming MCS tasks. The FL-based framework demonstrates not only benefits of illegitimate tasks detection, but also promising detection performance.  Our numerical results show that the FL system ensures promising improvements over earlier works \cite{chen2020deep} and can even achieve 100\% detection accuracy under chosen learning parameters.

The remainder of the paper is organized as follows. Section \ref{sec:relatedwork} discusses related work and its intent. Section \ref{sec:proposal} presents the FL-based detection system augmented with risk/loss-aware decision, dataset introduction and task definition. Section \ref{sec:numresults} shows experiments and numerical results. Section \ref{sec:conclusion} concludes the paper and highlights future research directions.

\section{Related Work and Motivation}
\label{sec:relatedwork}
Due to the MCS openness and opportunistic or participatory design of the data acquisition in MCS, sensed data collection campaigns are prone to various threats \cite{boubiche2019mobile}. For instance, malicious users distribute malware and eavesdropping attacks to steal sensitive information and damage users' devices \cite{dunham2008mobile}. Moreover, malicious participants' providing incorrect or low quality sensing data results in disinformation on the MCS platform \cite{li2019deep} is another threat analyzed in the literature. ML-based methods are designed to tackle various threats in MCS. For instance, a Q-learning and convolutional neural network-based method was proposed for malware identification \cite{wan2017reinforcement}. In order to detect illegitimate tasks, self-organizing feature map-based models are implemented in \cite{Zhang.2020.iot, chen2020locally}. 
 
FL-based approaches are introduced to address issues in MCS settings as FL is a promising AI technique that trains an algorithm across multiple decentralized edge devices or servers holding local data samples, without exchanging them \cite{yang2019federated}. The studies in \cite{kang2020reliable,wang.2020} aim to determine trusted and respectable users in FL tasks according to reputation metrics. In \cite{mowla2019federated}, the authors demonstrate a FL-backed approach to identify jamming attacks in flying ad-hoc networks and achieve a promising detection performance.

Motivated by the state of the art in fake task detection and the merit of  FL techniques, in this paper, to the best of our knowledge, for the first time we propose a FL-based framework to identify and mitigate illegitimate tasks in MCS platforms. 
\begin{figure}[h]
        \centering
        \includegraphics[width = 0.42\textwidth, trim=1.48cm 1.5cm 1.28cm 1cm,clip]{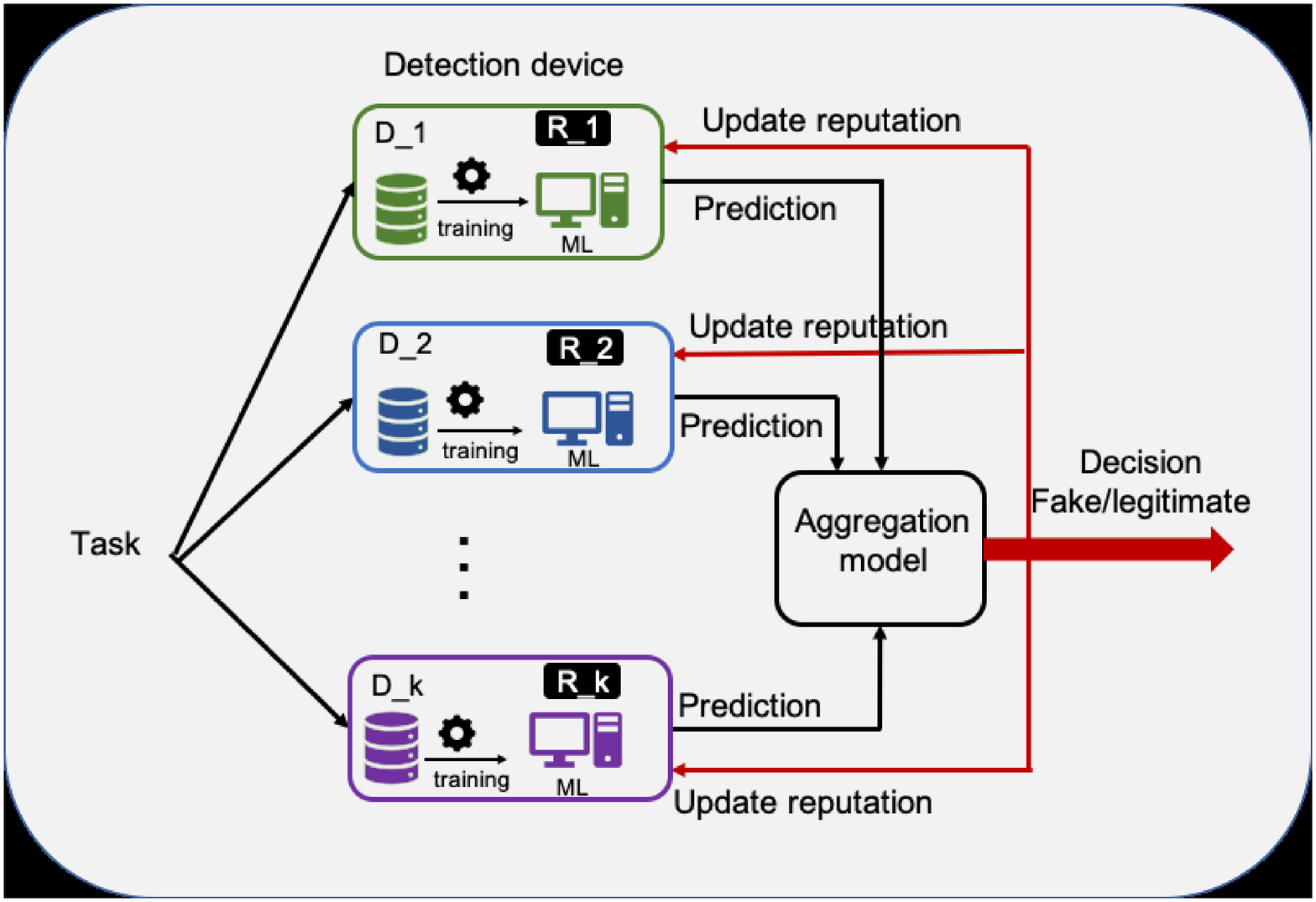}
        \caption{Federated learning based fake task detection system}
        \label{fig:mcs_systemoverview}
\end{figure}{}
\section{System Overview}
\label{sec:proposal}
Typically, MCS comprises three entities: service provider, end-users, and sensing data contributors (participants) \cite{wang2019coverage}. The service provider is responsible for implementing an MCS platform that connects with all three entities. End users initiate and transfer sensing tasks to the MCS platform while the MCS application center generates a task report to the users. Meanwhile, recruitment of participants, task distribution, sensing data collection, and rewards payment occur between the MCS application center and participants. However, non-dedicated and openness of MCS platforms lead to attackers submitting fake tasks to the platform in order to drain resources from participants’ devices and/or clog MCS servers, that's why detecting and mitigating illegitimate tasks from the MCS platform are essential to protect sensing data contributors.

\begin{table}[!hbt]
\centering
\caption{Simulation settings for task generation}
\label{tab:simulation}
\begin{tabular}{|p{1.5cm}|p{2.5cm}|p{3cm}|}
    \hline
    {}& Fake Tasks&Legitimate Tasks\\
    \hline
    Day&\multicolumn{2}{c|}{Distribution consistent in [1, 6]}\\
    \hline
    Hour &80\%: 7am-11am;
    20\%: 12pam-5pm& 8\%:0pm-5am; 92\%: 6am-23pm\\
    \hline
    Duration (minutes) &70\% in \{40, 50, 60\}; 
    30\% in \{10, 20, 30\} &Uniformly distributed over \{10, 20, 30, 40, 50, 60\}\\
    \hline
    Battery (\%) &80\% in \{7\%-10\%\};
    20\% in \{1\% -6\%\}& Distribution consistent in \{1\%-10\%\}\\
    \hline
    TaskValue &\multicolumn{2}{c|}{Uniformly distributed in [1,10]}\\
    \hline
    \end{tabular}
\end{table}

\subsection{Task definition and simulation settings }
Tasks are grouped into fake and legitimate categories. Fake tasks have two main objectives: 1) Higher battery drain from users' devices and 2) clogging MCS sensing server. Therefore, fake tasks are crucial threats affecting both an MCS platform and users. A task definition is introduced in \cite{audrey.sose2019} that consists of features \{'ID', 'latitude', 'longitude', 'day', 'hour', 'minute', 'duration', 'remaining time', 'battery requirement \%', 'Coverage', 'legitimacy', 'GridNumber', 'OnpeakHour'\}. Attribute 'ID' indicates task number that is not used by the ML model. Feature 'latitude' and 'longitude' determine the location of each task. Attribute 'GridNumber' is extracted from the sensing area map splitting. Feature 'day', 'hour', and 'minute' illustrate a task publishing time. Meanwhile, 'OnPeakHour' shows that the task runs during peak hours.  Feature 'duration' denotes the total time needed to complete this task. As to 'remaining time', it outlines the remaining time to finish a sensing task. A task describes the requirement of battery percentage in users' devices to complete a task. Feature 'coverage' describes the broadcast distance of tasks apart from distribution location. A task is labeled as legitimate or fake with respect to attribute 'legitimacy'. Moreover, we add a new feature, 'TaskValue', for tasks assigned an integer value uniformly from the range 1 to 10. 'TaskValue' is used for calculating loss rather than training ML models. As 'TaskValue' is a feature of a sample point, it can be obtained by task 'ID'. Simulation settings regarding task generation in simulations are given in Table ref{\tab:simulation}.

\begin{table} [!t]
\caption{Notation used in the paper{\vspace{-0.1in}}}
\centering
\label{tab:notations}
\begin{tabular}{|p{1cm}|p{7cm}|}
\hline
Notations & Description  \\\hline
$k$     & Total number of detection devices\\
$N$     & Total number of sensing tasks\\
$T_{s}$ &  Sets of tasks submitted by end users\\
$T_{i}$ & Task with index $i$ \\
$D_{j}$ & Dataset stored in $device_{i}$ \\\hline
$R_{i}^j$ & Reputation of  $device_{j}$ after predicting $task_{i}$ legitimacy \\
$R_{0}^j$ & Initial reputation of $device_{j}$ \\
$ES_{i}^j$ & ML algorithm in $device_{j}$ prediction value for $task_{i}$ \\
$DS_{i}^j$ & Prediction of $Device_j$ for $task_i$ legitimacy \\
$TV_{i}$ & Value of $task_{i}$\\\hline
$P(l|T_{i})$ & Probability of predicting $task_{i}$ as legitimate \\
$P(f|T_{i})$ & Probability of predicting $task_{i}$ as fake\\
$\lambda_{1}$ &  Incurred loss due to predicting a legitimate task as fake \\
$\lambda\_{2}$ &  Incurred loss due to predicting a fake task as legitimate \\
$RT_{i}$ & Risk of predicting $task_{i}$ as legitimate\\\hline
$RF_{i}$ & Risk of predicting $task_{i}$ as fake\\
$S_{i}$ & Aggregation results for $task_{i}$\\
$CP_{i}^j$ & Parameter to calculate reputation of $device_{j}$ after $task_{i}$\\
$CD_{i}^j$ &  Parameter to calculate reputation of $device_{j}$ after  $task_{i}$\\
$\epsilon$  & Extreme value for device reputation calculation\\\hline
$FD(T_{i})$  & Decision for $task_{i}$ made by aggregation module\\
$\alpha$  & Constant value\\
$DT\_0$ & Training dataset in \cite{chen2020deep} $DT\_0$ = $DT\_1$ $\cup$ $DT\_2$\\
$DT\_1$ & Half of $DT\_0$\\
$DT\_2$ & Half of $DT\_0$ and equal $(DT\_0$ - $DT\_1)$\\
\hline
\end{tabular}
\end{table}

\subsection{Reputation-aware Federated Learning Model}
\label{sec:rep_sys}


To explain the proposed methodology, notations in the mathematical formulas are gathered in Table \ref{tab:notations}. The proposed FL-based detection system deploys several distributed devices with a local dataset and a ML model, as illustrated in Fig.\ref{fig:mcs_systemoverview}. Each device aggregates sensing tasks upfront and stores the dataset locally. The ML model in each device is trained using a local dataset before receiving a new sensing task. The MCS platform then distributes tasks submitted from end-users to all detection devices. The trained ML models in devices estimate a task label. In an aggregation module, a risk-aware method is utilized to quantify the loss of all possible actions in the decision-making stage. Before introducing risk-awareness, the initial aggregated legitimacy decision for $task_{i}$ ($S_{i}$) is calculated as $ S_{i} = \sum_{j=1}^{k} DS_{i}^j$ where the decision of $device_{j}$ for $task_{i}$ ($DS_{i}^j$) is weighted by the reputation of the device at the end of the previous aggregated decision for $task_{i-1}$ as in (\ref{equ:device_decision}) where $ES_{i}^j$ stands for the local ML model-based prediction value of $device_{j}$, and $R_{i-1}^j$ is $device_{j}$ reputation after $task_{i-1}$. We define fake tasks and legitimate tasks with the labels $0$ and  $1$, respectively. Based on  (\ref{equ:device_decision}), the initial aggregated legitimacy can be re-formulated as  in (\ref{equ:aggr_2}).



\begin{equation}
    \label{equ:device_decision}
    DS_{i}^j = ES_{i}^j  R_{i-1}^j
\end{equation}
\vspace{-0.2in}
 \begin{equation}
\label{equ:aggr_2}
    S_{i} = \sum_{j=1}^{k} ES_{i}^j R_{i-1}^j
\end{equation}
Device reputation is updated upon every aggregated decision on task legitimacy. Reputation of $device_{j}$ after $task_i$ is calculated as in (\ref{equ:reputaion}) where $CP_{i}^j$ and $CD_{i}^j$ are used to adjust the reputation of $device_{j}$. The two parameters start with $0$ so devices are assigned an initial reputation of 0.5. If the local ML model of $device_{j}$ outputs the same result as the aggregation module, both  $CP_{i}^j$ and $CD_{i}^j$ are incremented by one. Otherwise, the value of $CP_{i}^j$ remains while $CD_{i}^j$ is incremented. Increasing $CP_{i}^j$ will favor the reputation of $device_{j}$, whereas an increase in $CD_{i}^j$ could drag reputation down. In an ideal situation, the decision of $device_{j}$ always matches the aggregated decision, and after $N$ task decisions, the values of $CP_{N}^j$ and $CP_{N}^j$ reach $N-1$. 

\begin{equation}
\label{equ:reputaion}
\begin{aligned}
  R_{i}^j =
    \begin{cases}
      \frac{\epsilon+CP_{i}^j}{2\times \epsilon+CD_{i}^j} & i \geq 2\\
      0.5 & i=1, CP_{i}^j = CD_{i}^j = 0\\
    \end{cases}       
\end{aligned}
\end{equation}
 According to (\ref{equ:reputaion}), reputation is going to be calculated as follows for an always correct device:

\begin{equation}
    \label{equ:reputation_N}
    R_{N}^j = \frac{\epsilon+(N-1)}{2\times \epsilon+(N-1)}
\end{equation}{}

Since $\epsilon$ is a negligibly small value, the limit of   (\ref{equ:reputation_N}) approaches 1 as expressed in   (\ref{equ:reputation_limit}).
\begin{equation}
    \label{equ:reputation_limit}
    \lim_{N\to\infty} R_{N}^j=\lim_{N\to\infty} \frac{\epsilon+(N-1)}{2\times \epsilon+(N-1)} = 1
\end{equation}{}
Thus, reputation values for all devices should be less than $1$ due to the limitation of total task number $N$ and almost impossible $100\%$ matching with aggregation module. According to  (\ref{equ:aggr_2}) and (\ref{equ:reputaion}), the reputation-aware aggregation values can be formulated by (\ref{equ:aggr_3}).

\begin{equation}
\label{equ:aggr_3}
\begin{aligned}
  S_{i} =
    \begin{cases}
      \sum_{j=1}^{k} ES_{i}^j \frac{\epsilon+CP_{i}^j}{2\times \epsilon+CD_{i}^j} & i \geq 2\\
     \frac{1}{2} \sum_{j=1}^{k} ES_{i}^j & i=1 \\
    \end{cases}       
\end{aligned}
\end{equation}
Equation  (\ref{equ:aggr_3}) defines the aggregation results of $task_{i}$ that relies on the ML-based estimation value $ES_{i}^j$ and the device reputation. As introduced before, $ES_{i}^j$ is either $0$ or $1$ representing malicious tasks and legitimate tasks, respectively. 
If $device_{j}$ estimates $task_{i}$ as fake, $ES_{i}^j$ is set to $0$. In this case, devices predicting $task_{i}$ as illegitimate do not contribute in updating $S_{i}$. Accordingly, (\ref{equ:aggr_3}) formulates a summary of legitimate prediction for $task_{i}$. Based on these analyses, the overall probability of task prediction to be legitimate is formulated by (\ref{equ:probability_leg}).
\begin{equation}
    \label{equ:probability_leg}
    P(l|T_{i}) = \frac{S_{i}}{k}
\end{equation}{}
 $P(l|T_{i})$ represents probability of prediction $task_{i}$ as legitimate. Meanwhile, $P(l|T_{i})$ and $P(f|T_{i})$ should satisfy the condition: $ P(l|T_{i}) + P(f|T_{i}) = 1$. 
According to this condition and (\ref{equ:probability_leg}), probability of predicting $task_{i}$ as fake is obtained in (\ref{equ:probability_fake}).
\begin{equation}
    \label{equ:probability_fake}
    P(f|T_{i}) = 1 - P(l|T_{i})
\end{equation}{}


\begin{table} [!hbt]
\caption{Utility loss description\vspace{-0.2in}}
\label{tab:loss_matric}
\center
\begin{tabular}{|p{2cm}|p{2cm}|p{2cm}|}
\hline
&Legitimate task & Fake task  \\\hline
Predict legitimate& 0   & $\lambda_{1}$ \\
Predict fake & $\lambda_{2}$ & 0\\
\hline
\end{tabular}
\end{table}

Table \ref{tab:loss_matric} presents the utility loss due to classification decisions. Specifically, the ML model predicting a fake task as fake or predicting a legitimate task as legitimate impacts the MCS platform's utility. On the other hand, incorrect estimation results in a loss. Thus, $\lambda_{1}$ is the loss due to classifying a legitimate task as fake, while $\lambda_{2}$ is the loss incurred due to classifying a fake task as legitimate. Therefore, each decision action entails a certain risk, and the risks of legitimate and fake predictions are formulated respectively as follows:
\begin{equation}
    \label{equ:loss_leg}
    RT_{i} =  P(f|T_{i}) \times \lambda_{1} \times TV_{i}
\end{equation}{}
\begin{equation}
    \label{equ:loss_fake}
    RF_{i} =  P(l|T_{i}) \times \lambda_{2}\times TV_{i}
\end{equation}{}
Equation (\ref{equ:loss_leg}) define the risk of predicting $task_{i}$ as legitimate and Equation (\ref{equ:loss_fake}) formulates the risk of predicting $task_{i}$ as fake. Aggregation module determines the legitimacy of a task according to a risk value. Specifically, risk is a function of utility loss due to a taken action as formulated in  (\ref{equ:decision_aggr}) where $FD(T_{i})$ denotes the final aggregated decision for $task_{i}$. Thus, the final decision chooses the least risky action.

\begin{equation}
\label{equ:decision_aggr}
\begin{aligned}
  FD(T_{i}) =
    \begin{cases}
      1 & RT_{i}\leq RF_{i} \\
      0 & RT_{i}> RF_{i}\\
    \end{cases}       
\end{aligned}
\end{equation}

\subsection{Vote based federated learning system}
With the concept of static reputation value for devices, we design a vote-based aggregation approach in the FL platform as an alternative to the reputation-based approach. Thus, $R_{i}^j$ is set to a constant value of $\alpha$ that remains the same for all tasks. Then aggregation value  in (\ref{equ:aggr_2}) can be re-formulated as follows:
\begin{equation}
\label{equ:aggr_vote}
    S_{i} = \sum_{j=1}^{k} ES_{i}^j \times \alpha
\end{equation}

As introduced in Section \ref{sec:rep_sys}, a device's reputation should not exceed $1$. As a result, $\alpha$ should be in the range of $0$ to $1$. The device's decision is eliminated by the aggregation module setting $\alpha=0$, whereas $\alpha=1$ stands for completely accepting the device's decision based upon (\ref{equ:aggr_vote}). Meanwhile, devices are partially trusted when $\alpha$ is less than $1$. To reduce the computational complexity, $\alpha$ is chosen as $1$.
Probability of predicting $task_{i}$ as legitimate is formulated as shown in (\ref{equ:prob_leg_vote}):
\begin{equation}
\label{equ:prob_leg_vote}
    P(l|T_{i}) = \frac{\sum_{j=1}^{k} ES_{i}^j}{k}
\end{equation}{}

According to (\ref{equ:probability_fake}) and (\ref{equ:prob_leg_vote}), $P(f|T_{i})$ is deduced. 
Loss calculation and aggregation decision rules are the same as reputation-aware FL model showing (\ref{equ:loss_leg}),   (\ref{equ:loss_fake}) and (\ref{equ:decision_aggr}), respectively.

\section{Performance Evaluation}
\label{sec:numresults}


Based on the design of tasks, the CrowdSenSim tool generates the dataset using real physical features (e.g., using Timmins, a small city in Canada as a testbed \cite{CrowdSensim} . It is an imbalanced dataset with $89\%$ legitimate tasks and $11\%$ fake tasks. In this paper, the dataset is composed of $1,000$ MCS tasks in total, as noted in our prior work  \cite{chen2020deep}. In \cite{chen2020deep}, $800$ tasks are used for training a DBN model and $200$ tasks are used for testing. In Table \ref{tab:notations}, $DT\_0$ contains $800$ tasks that is the same as the training dataset in \cite{chen2020deep}. Here we keep the same test dataset to compare fairly with previous results. 


Comprehensive experiments are executed to verify performance of mitigating illegitimate tasks based on different configurations including 1) reputation (e.g., dynamic and static), 2) loss metric (e.g., $\lambda_{1}$ and  $\lambda_{2}$), and 3) the size of the local dataset saved in devices (e.g., $800$ tasks and $400$ tasks). Meanwhile, we compare proposed  FL performance with centralized system detection results performed by one ML model. Considering system complexity, the number of detection devices $k$ should not be a large value. In this paper,  $k$ is set up as $5$. Meanwhile, we set $\epsilon$ at $(1.0e-5)$. 
\begin{table}[!t]
\caption{Performance by centralized system}
\label{tab:per_cent}
\centering
\begin{tabular}{ccccccllll}
\cline{1-6}
\multicolumn{1}{|c|}{Device} & \multicolumn{1}{c|}{Precision} & \multicolumn{1}{c|}{Recall} & \multicolumn{1}{c|}{F1}     & \multicolumn{1}{c|}{G-mean} & \multicolumn{1}{c|}{Accuracy} &  &  &  &  \\ \cline{1-6}
\multicolumn{1}{|c|}{1}      & \multicolumn{1}{c|}{0.9848}    & \multicolumn{1}{c|}{0.9850} & \multicolumn{1}{c|}{0.9848} & \multicolumn{1}{c|}{0.9511} & \multicolumn{1}{c|}{0.9850}   &  &  &  &  \\ \cline{1-6}
\multicolumn{1}{|c|}{2}      & \multicolumn{1}{c|}{0.9900}    & \multicolumn{1}{c|}{0.9900} & \multicolumn{1}{c|}{0.9900} & \multicolumn{1}{c|}{0.9743} & \multicolumn{1}{c|}{0.9900}   &  &  &  &  \\ \cline{1-6}
\multicolumn{1}{|c|}{3}      & \multicolumn{1}{c|}{0.9901}    & \multicolumn{1}{c|}{0.9901} & \multicolumn{1}{c|}{0.9898} & \multicolumn{1}{c|}{0.9539} & \multicolumn{1}{c|}{0.9900}   &  &  &  &  \\ \cline{1-6}
\multicolumn{1}{|c|}{4}      & \multicolumn{1}{c|}{0.9852}    & \multicolumn{1}{c|}{0.9850} & \multicolumn{1}{c|}{0.9845} & \multicolumn{1}{c|}{0.9303} & \multicolumn{1}{c|}{0.9850}   &  &  &  &  \\ \cline{1-6}
\multicolumn{1}{|c|}{5}      & \multicolumn{1}{c|}{0.9900}    & \multicolumn{1}{c|}{0.9900} & \multicolumn{1}{c|}{0.9900} & \multicolumn{1}{c|}{0.9743} & \multicolumn{1}{c|}{0.9900}   &  &  &  &  \\ \cline{1-6}
\multicolumn{1}{|c|}{DBN \cite{chen2020deep}}    & \multicolumn{1}{c|}{0.9420}    & \multicolumn{1}{c|}{0.9230} & \multicolumn{1}{c|}{0.9280} & \multicolumn{1}{c|}{0.8697} & \multicolumn{1}{c|}{0.9230}   &  &  &  &  \\ \cline{1-6}
\multicolumn{1}{l}{}         & \multicolumn{1}{l}{}           & \multicolumn{1}{l}{}        & \multicolumn{1}{l}{}        & \multicolumn{1}{l}{}        & \multicolumn{1}{l}{}          &  &  &  &  \\
\multicolumn{1}{l}{}         & \multicolumn{1}{l}{}           & \multicolumn{1}{l}{}        & \multicolumn{1}{l}{}        & \multicolumn{1}{l}{}        & \multicolumn{1}{l}{}          &  &  &  & 
\end{tabular}
\vspace{-0.3in}
\end{table}

\begin{figure}[!t]
        \centering
        \includegraphics[width = 0.5\textwidth, trim=3cm 3cm 0cm 3cm,clip]{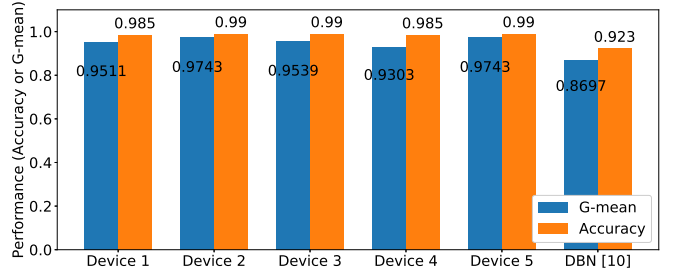}
        \caption{Centralized performance 
        }
        \label{fig:centralized_res}
\end{figure}

\subsection{Centralized test platform performance}
The study in \cite{chen2020deep} demonstrated a Deep Belief Network (DBN)-based fake task detection system in MCS. In this paper, several ensemble algorithms such as XGBoost \cite{chen2015xgboost}, AdaBoost\cite{schapire2013explaining} and Bagging \cite{breiman1996bagging} are used to mitigate malicious tasks. In centralized case, we can consider the aggregation module is disabled in the proposed system Fig. \ref{fig:mcs_systemoverview}. The centralized test model configuration is shown in Table \ref{tab:cen_config}. The training dataset and test dataset are the same as in \cite{chen2020deep}. Table \ref{tab:per_cent} presents prediction results. DBN results is referred in \cite{chen2020deep}. According to the results, all ensemble models boost detection performance dramatically than DBN \cite{chen2020deep}. Fig.\ref{fig:centralized_res} illustrates G-mean and accuracy comparison results. It shows XGBoost in device $5$ and Adaboost in device $2$ demonstrate the highest G-mean  achieving over $12\%$ improvement over DBN.

\subsection{Federated learning-based system}
\begin{figure}[!t]
        \centering
        \includegraphics[width = 0.5\textwidth, trim=3cm 2.3cm 2cm 3.5cm,clip]{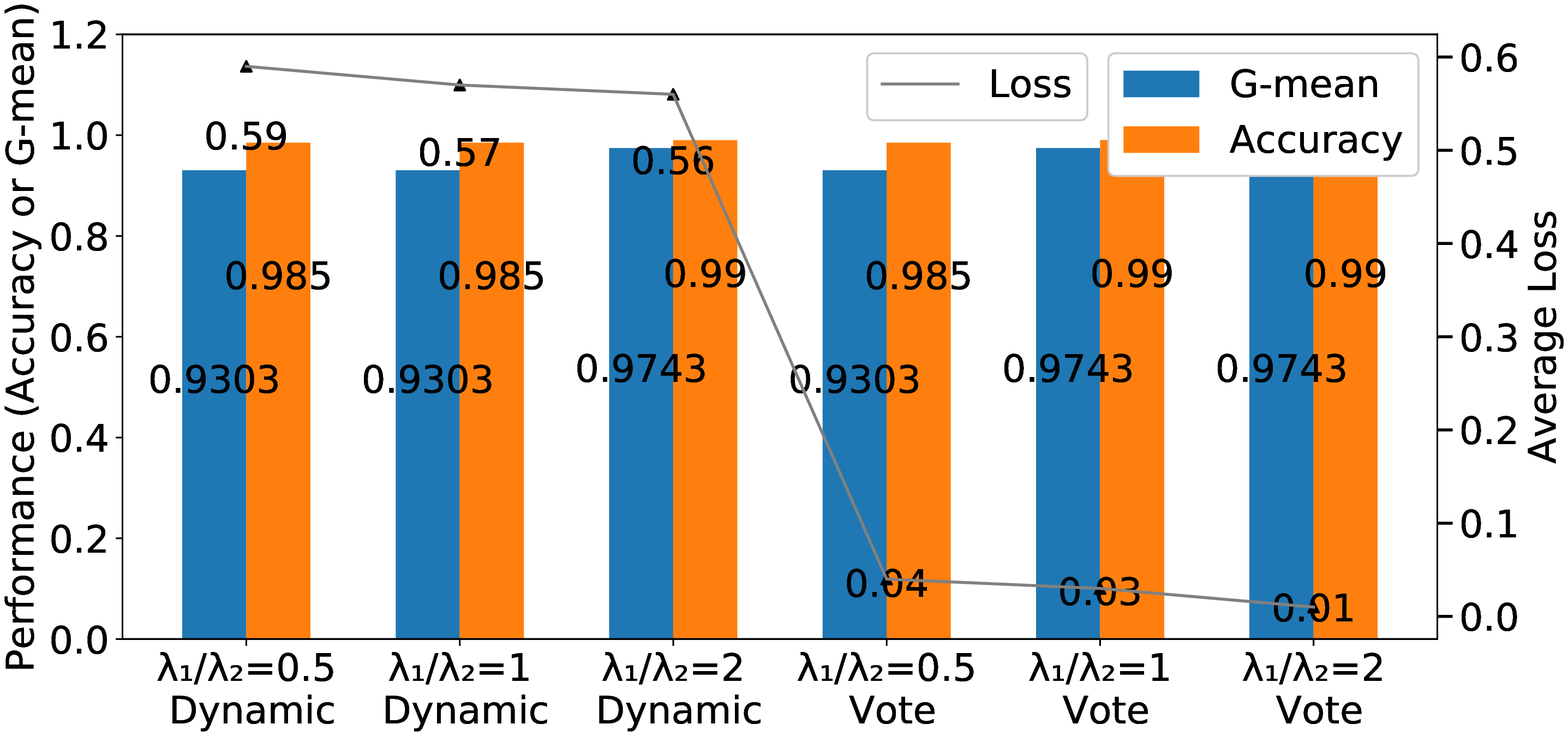}
        \caption{FL-based performance comparison in MODEL\_1\vspace{-0.1in}}
        \label{fig:fed_learn_large}
        
\end{figure}

Through deploying different machine learning models and different local datasets in devices, we design two FL-based models. FL MODEL\_1 configuration is the same as centralized system in Table  \ref{tab:cen_config}. In this model, $800$ samples are saved in all devices in advance. The FL MODEL\_2 configuration shows in Table \ref{tab:fed_model_conf}. In this model, only $400$ data points are distributed to devices. Loss parameters $\lambda_{1}$ and $\lambda_{2}$ in Table\ref{tab:loss_matric} are configured three different groups: (1)$\lambda_{1}$=1, $\lambda_{2}$=2; (2)$\lambda_{1}$=1, $\lambda_{2}$=1, and (3)$\lambda_{1}$=1, $\lambda_{2}$=0.5. Furthermore, both reputation aware and vote base approach with static reputation are considered. Table \ref{tab:feder_large_perform} presents MODEL\_1 performance and average loss value. Table \ref{tab:feder_large_perform} shows the best detection performance in group (3) ($\lambda_{1}$=1, $\lambda_{2}$=0.5) for both dynamic and static reputation cases with up to $0.99$ accuracy. From increasing the value of $\lambda_{2}$ boosts platform performance as well as marginally decreases average loss for both dynamic reputation and vote-based cases. Specifically, $\lambda_{2}$ increasing from 0.5 to 2 in dynamic reputation case, gives G-mean value improvement from $0.9303$ to $0.9743$ and loss per device decreases from 0.59 to 0.56. In the vote based case, the trend of loss and performance is the same as the dynamic reputation case. Raising $\lambda_{2}$ boosts detection accuracy and lowers loss. Fig.\ref{fig:fed_learn_large} illustrates accuracy and G-mean comparison with different $\lambda_{2}$ values and  reputation determination rules (i.e., vote-based or dynamic assessment) when $800$ samples are fed in detection devices. 

\begin{table}[!t]
\centering
\caption{Centralized model and FL-based model (MODEL\_1) configuration. Num of BES denotes the number of base estimators.}
\label{tab:cen_config}
\begin{tabular}{ccccclllll}
\cline{1-5}

\multicolumn{1}{|c|}{Device} & \multicolumn{1}{c|}{ML}       & \multicolumn{1}{c|}{Base estimator}        & \multicolumn{1}{c|}{Num of BES} & \multicolumn{1}{c|}{Dataset} &  &  &  &  &  \\ \cline{1-5}
\multicolumn{1}{|c|}{1}      & \multicolumn{1}{c|}{Bagging}  & \multicolumn{1}{c|}{Xgboost}               & \multicolumn{1}{c|}{100}               & \multicolumn{1}{c|}{DT\_0}   &  &  &  &  &  \\ \cline{1-5}
\multicolumn{1}{|c|}{2}      & \multicolumn{1}{c|}{Adaboost} & \multicolumn{1}{c|}{DecisionTreeRegressor} & \multicolumn{1}{c|}{50}              & \multicolumn{1}{c|}{DT\_0}   &  &  &  &  &  \\ \cline{1-5}
\multicolumn{1}{|c|}{3}      & \multicolumn{1}{c|}{Bagging}  & \multicolumn{1}{c|}{DecisionTree}          & \multicolumn{1}{c|}{100}              & \multicolumn{1}{c|}{DT\_0}   &  &  &  &  &  \\ \cline{1-5}
\multicolumn{1}{|c|}{4}      & \multicolumn{1}{c|}{Bagging}  & \multicolumn{1}{c|}{RandomForest}          & \multicolumn{1}{c|}{100}              & \multicolumn{1}{c|}{DT\_0}   &  &  &  &  &  \\ \cline{1-5}
\multicolumn{1}{|c|}{5}      & \multicolumn{1}{c|}{XGBoost}  & \multicolumn{1}{c|}{N/A}                   & \multicolumn{1}{c|}{50}               & \multicolumn{1}{c|}{DT\_0}   &  &  &  &  &  \\ \cline{1-5}
\multicolumn{1}{l}{}         & \multicolumn{1}{l}{}          & \multicolumn{1}{l}{}                       & \multicolumn{1}{l}{}                  & \multicolumn{1}{l}{}         &  &  &  &  &  \\
\multicolumn{1}{l}{}         & \multicolumn{1}{l}{}          & \multicolumn{1}{l}{}                       & \multicolumn{1}{l}{}                  & \multicolumn{1}{l}{}         &  &  &  &  &  \\
\multicolumn{1}{l}{}         & \multicolumn{1}{l}{}          & \multicolumn{1}{l}{}                       & \multicolumn{1}{l}{}                  & \multicolumn{1}{l}{}         &  &  &  &  & 
\end{tabular}
\vspace{-0.4in}
\end{table}

\begin{table}[!t]
\centering
\caption{Federated learning MODEL\_2 }
\label{tab:fed_model_conf}
\begin{tabular}{ccccclllll}

\cline{1-5}
\multicolumn{1}{|c|}{Device} & \multicolumn{1}{c|}{ML}       & \multicolumn{1}{c|}{Base estimator}        & \multicolumn{1}{c|}{Num of BES} & \multicolumn{1}{c|}{Dataset} &  &  &  &  &  \\ \cline{1-5}
\multicolumn{1}{|c|}{1}      & \multicolumn{1}{c|}{Adaboost} & \multicolumn{1}{c|}{DecisionTreeRegressor} & \multicolumn{1}{c|}{100}        & \multicolumn{1}{c|}{DT\_1}   &  &  &  &  &  \\ \cline{1-5}
\multicolumn{1}{|c|}{2}      & \multicolumn{1}{c|}{Bagging}  & \multicolumn{1}{c|}{DecisionTree}          & \multicolumn{1}{c|}{50}         & \multicolumn{1}{c|}{DT\_1}   &  &  &  &  &  \\ \cline{1-5}
\multicolumn{1}{|c|}{3}      & \multicolumn{1}{c|}{Bagging}  & \multicolumn{1}{c|}{DecisionTree}          & \multicolumn{1}{c|}{100}        & \multicolumn{1}{c|}{DT\_2}   &  &  &  &  &  \\ \cline{1-5}
\multicolumn{1}{|c|}{4}      & \multicolumn{1}{c|}{XGBoost}  & \multicolumn{1}{c|}{N/A}                   & \multicolumn{1}{c|}{100}        & \multicolumn{1}{c|}{DT\_1}   &  &  &  &  &  \\ \cline{1-5}
\multicolumn{1}{|c|}{5}      & \multicolumn{1}{c|}{XGBoost}  & \multicolumn{1}{c|}{N/A}                   & \multicolumn{1}{c|}{50}         & \multicolumn{1}{c|}{DT\_2}   &  &  &  &  &  \\ \cline{1-5}
\multicolumn{1}{l}{}         & \multicolumn{1}{l}{}          & \multicolumn{1}{l}{}                       & \multicolumn{1}{l}{}            & \multicolumn{1}{l}{}         &  &  &  &  &  \\
\multicolumn{1}{l}{}         & \multicolumn{1}{l}{}          & \multicolumn{1}{l}{}                       & \multicolumn{1}{l}{}            & \multicolumn{1}{l}{}         &  &  &  &  &  \\
\multicolumn{1}{l}{}         & \multicolumn{1}{l}{}          & \multicolumn{1}{l}{}                       & \multicolumn{1}{l}{}            & \multicolumn{1}{l}{}         &  &  &  &  & 
\end{tabular}
\vspace{-0.3in}
\end{table}

\begin{figure}[!t]
\center
    \begin{subfigure}{0.40\textwidth}
        \centering
        \includegraphics[width = 1\textwidth, trim=2cm 0cm 1.5cm 2.2cm,clip]{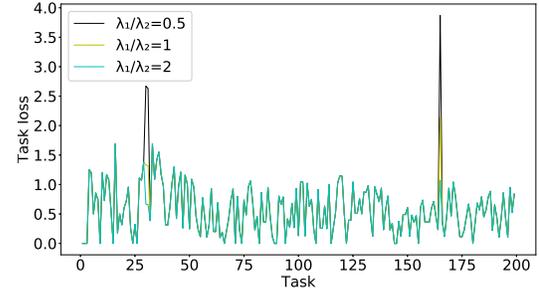}
        \caption{Task loss with dynamic reputation}
        \label{fig:loss_task_rel_rep}
    \end{subfigure}
    \begin{subfigure}{0.40\textwidth}
        \centering
        \includegraphics[width = 1\textwidth, trim=2cm 0cm 1.5cm 2.2cm,clip]{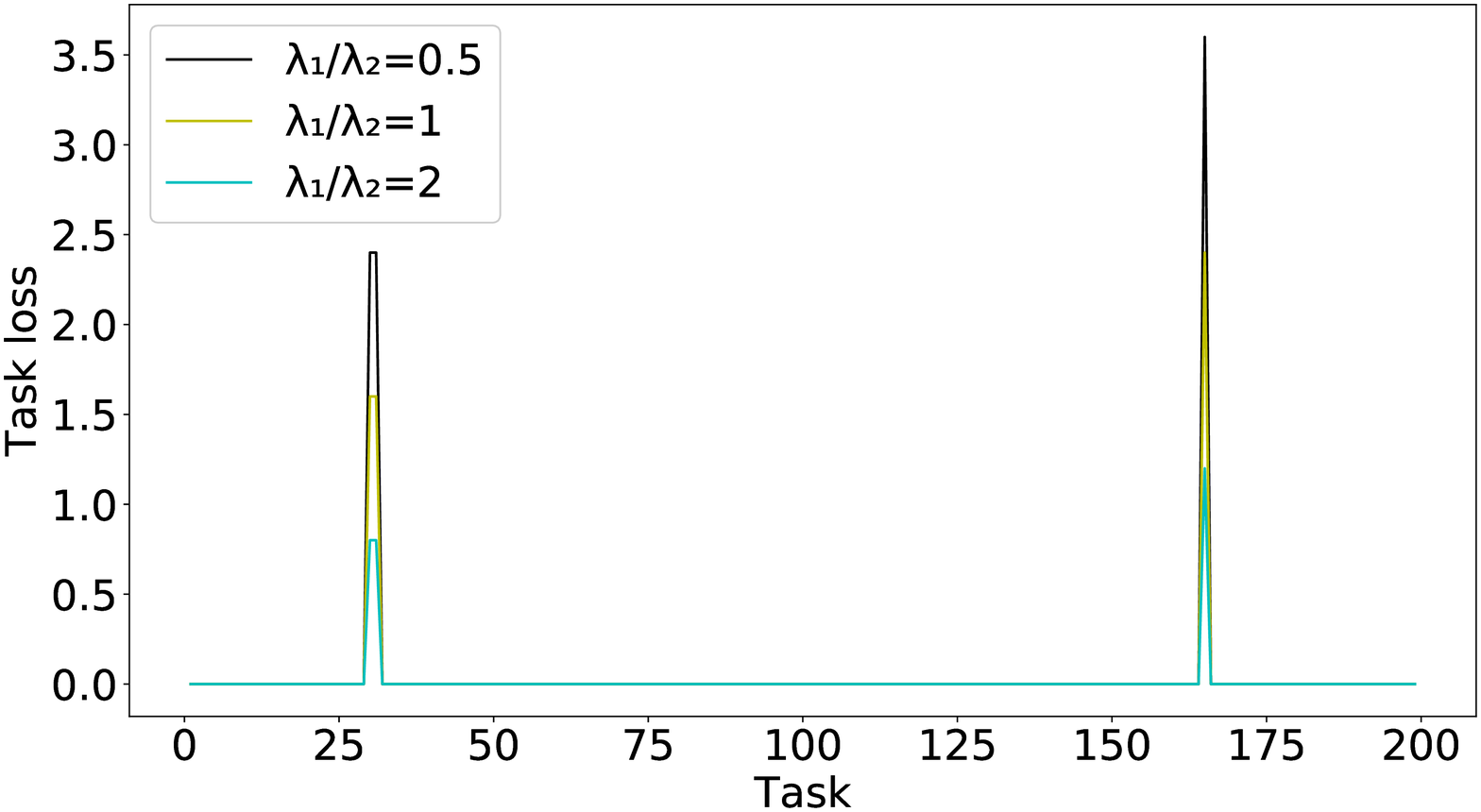}
        \caption{Task loss with static reputation}
        \label{fig:loss_task_rel_vote}
    \end{subfigure}
    
    \caption{Task value loss under dynamic reputation and static reputation in MODEL\_1 \vspace{-0.1in} }
	\label{fig:alltaskloss}
\end{figure}

Results in Table \ref{tab:feder_large_perform} describe average loss in static is critically lower than dynamic. The root cause is that the reputation is always less than "1" based on   (\ref{equ:reputation_limit}) if $N$ is set to 200. It results in aggregation value $S_{i}$ for $task_{i}$ in (\ref{equ:aggr_3}) should be less than $5$ even if all devices predict this task as legitimate (5 devices model designed here). Therefore, probability of prediction task as legitimate in (\ref{equ:probability_leg}) should be always less than $1$ and $P(f|T_{i})$ always a positive value according to   (\ref{equ:probability_fake}). It means prediction risk $RT_{i}$ and $RF_{i}$ are always a positive value. On the other hand, vote based system with static reputation 1 avoids loss when five devices estimate $task_{i}$ as legitimate. In this case, aggregation $S_{i}$ is $5$ according to (\ref{equ:aggr_vote}). We can get $P(l|T_{i})$ as 1 in (\ref{equ:prob_leg_vote}) and $P(f|T_{i})$ as 0 in (\ref{equ:probability_fake}). With $P(f|T_{i})$ 0, $RT_{i}$ in (\ref{equ:loss_leg}) is $0$. Fig.\ref{fig:alltaskloss} demonstrates loss of $200$ test tasks with dynamic and static reputation respectively based on different loss metric parameters. From Fig. \ref{fig:alltaskloss}, most tasks loss in vote based MODEL\_2 is $0$ that contributes a smaller average loss than reputation aware MODEL\_1.

\begin{figure}[!t]
        \centering
        \includegraphics[width = 0.5\textwidth, trim=3cm 2.3cm 2cm 3.5cm,clip]{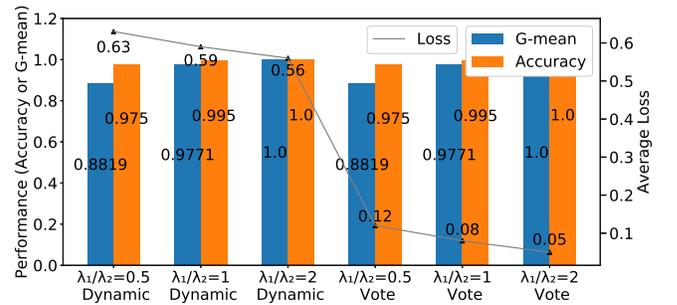}
        \caption{FL-based performance comparison in MODEL\_2}
        \label{fig:fed_learn_small}
        \vspace{-0.5cm}
\end{figure}

Table \ref{tab:feder_small_perform} describes MODEL\_2 performance and prediction loss. It shows the same trend as in MODEL\_1 in terms of suppressing rate of $\lambda_{1}$ over $\lambda_{2}$ resulting in a reduction of loss and improvement of detection performance. In the case ($\lambda_{1},\lambda_{2})= (1,2)$, we get the least loss and $100\%$ performance for both dynamic and static reputation considered cases. Two factors contribute to $100\%$ accuracy and precision. Firstly, individual detection device performs encouraging detection accuracy showing in centralized detection model in Table \ref{tab:per_cent}. Secondly, the FL-based method relies on horizontal device decisions and gives different decisions via configuring loss metric parameters. 
Estimation $DS_{i}^j$ of deployed five devices is 0, 0, 1, 1, 0 respectively. We then get aggregation result $S_{i}$ as 2. According to (\ref{equ:prob_leg_vote}), probability of legitimate is calculated as $0.4$. After that, loss of predicting task as legitimate and loss is measured according to (\ref{equ:loss_leg}) and (\ref{equ:loss_fake}) using different loss parameters $\lambda_{1}$ and $\lambda_{2}$. When $\lambda_{2}$=2 $\lambda_{1}$, the aggregation module determines $task_{i}$ legitimate that is incorrect decision comparing with task label. If $\lambda_{1}$/ $\lambda_{2}$ is either to 1 or 2, FL-based model decides $task_{i}$ fake that is correct estimation. Therefore, adjusting $\lambda_{1}$/ $\lambda_{2}$, aggregation makes different decision of a task.

Comparing results in MODEL\_1 (Table \ref{tab:feder_large_perform}) and MODEL\_2 (Table \ref{tab:feder_small_perform}), MODEL\_2 shows marginal improvement than MODEL\_1 when $\lambda_{1}$/$\lambda_{2}$ is chosen as $1$ and $2$. More specifically, G-mean shows over $2.6\%$ improvement from $0.9743$ in MODEL\_1 to $1$ in MODEL\_2 selecting $\lambda_{1}$/$\lambda_{2}$ as 2.

\begin{table} [!t]
\caption{Performance in FL system MODEL\_1\vspace{-0.15in}}
\center
\label{tab:feder_large_perform}
\begin{tabular}{ccccccclll}
\cline{1-7}
\multicolumn{1}{|c|}{Config}                                                 & \multicolumn{1}{c|}{Prec.} & \multicolumn{1}{c|}{Recall} & \multicolumn{1}{c|}{F1}     & \multicolumn{1}{c|}{G-mean} & \multicolumn{1}{c|}{Acc.} & \multicolumn{1}{c|}{loss} &  &  &  \\ \cline{1-7}
\multicolumn{1}{|c|}{\begin{tabular}[c]{@{}c@{}}$\lambda_{1}$/$\lambda_{2}$=0.5\\ Dynamic\end{tabular}} & \multicolumn{1}{c|}{0.9852}    & \multicolumn{1}{c|}{0.9850} & \multicolumn{1}{c|}{0.9845} & \multicolumn{1}{c|}{0.9303} & \multicolumn{1}{c|}{0.9850}   & \multicolumn{1}{c|}{0.59}        &  &  &  \\ \cline{1-7}
\multicolumn{1}{|c|}{\begin{tabular}[c]{@{}c@{}}$\lambda_{1}$/$\lambda_{2}$=1\\ Dynamic\end{tabular}}   & \multicolumn{1}{c|}{0.9852}    & \multicolumn{1}{c|}{0.9850} & \multicolumn{1}{c|}{0.9845} & \multicolumn{1}{c|}{0.9303} & \multicolumn{1}{c|}{0.9850}   & \multicolumn{1}{c|}{0.57}        &  &  &  \\ \cline{1-7}
\multicolumn{1}{|c|}{\begin{tabular}[c]{@{}c@{}}$\lambda_{1}$/$\lambda_{2}$=2\\ Dynamic\end{tabular}}   & \multicolumn{1}{c|}{0.9900}    & \multicolumn{1}{c|}{0.9900} & \multicolumn{1}{c|}{0.9900} & \multicolumn{1}{c|}{0.9743} & \multicolumn{1}{c|}{0.9900}   & \multicolumn{1}{c|}{0.56}        &  &  &  \\ \cline{1-7}
\multicolumn{1}{|c|}{\begin{tabular}[c]{@{}c@{}}$\lambda_{1}$/$\lambda_{2}$=0.5\\ Vote\end{tabular}}  & \multicolumn{1}{c|}{0.9852}    & \multicolumn{1}{c|}{0.985}  & \multicolumn{1}{c|}{0.9845} & \multicolumn{1}{c|}{0.9303} & \multicolumn{1}{c|}{0.985}    & \multicolumn{1}{c|}{0.04}        &  &  &  \\ \cline{1-7}
\multicolumn{1}{|c|}{\begin{tabular}[c]{@{}c@{}}$\lambda_{1}$/$\lambda_{2}$=1\\ Vote\end{tabular}}    & \multicolumn{1}{c|}{0.9900}    & \multicolumn{1}{c|}{0.9900} & \multicolumn{1}{c|}{0.9900} & \multicolumn{1}{c|}{0.9743} & \multicolumn{1}{c|}{0.9900}   & \multicolumn{1}{c|}{0.03}        &  &  &  \\ \cline{1-7}
\multicolumn{1}{|c|}{\begin{tabular}[c]{@{}c@{}}$\lambda_{1}$/$\lambda_{2}$=2\\ Vote\end{tabular}}    & \multicolumn{1}{c|}{0.9900}    & \multicolumn{1}{c|}{0.9900} & \multicolumn{1}{c|}{0.9900} & \multicolumn{1}{c|}{0.9743} & \multicolumn{1}{c|}{0.9900}   & \multicolumn{1}{c|}{0.01}        &  &  &  \\ \cline{1-7}
\multicolumn{1}{l}{}                                                                & \multicolumn{1}{l}{}           & \multicolumn{1}{l}{}        & \multicolumn{1}{l}{}        & \multicolumn{1}{l}{}        & \multicolumn{1}{l}{}          & \multicolumn{1}{l}{}             &  &  &  \\
\multicolumn{1}{l}{}                                                                & \multicolumn{1}{l}{}           & \multicolumn{1}{l}{}        & \multicolumn{1}{l}{}        & \multicolumn{1}{l}{}        & \multicolumn{1}{l}{}          & \multicolumn{1}{l}{}             &  &  & 
\end{tabular}
\vspace{-0.2in}
\end{table}

\begin{table}[!t]
\caption{Performance in FL system MODEL\_2\vspace{-0.1in}}
\centering
\label{tab:feder_small_perform}
\begin{tabular}{ccccccclll}

\cline{1-7}
\multicolumn{1}{|c|}{Config}                                                 & \multicolumn{1}{c|}{Prec.} & \multicolumn{1}{c|}{Recall} & \multicolumn{1}{c|}{F1}     & \multicolumn{1}{c|}{G-mean} & \multicolumn{1}{c|}{Acc} & \multicolumn{1}{c|}{loss} &  &  &  \\ \cline{1-7}
\multicolumn{1}{|c|}{\begin{tabular}[c]{@{}c@{}}$\lambda_{1}$/$\lambda_{2}$=0.5\\ Dynamic\end{tabular}} & \multicolumn{1}{c|}{0.9757}    & \multicolumn{1}{c|}{0.9750} & \multicolumn{1}{c|}{0.9736} & \multicolumn{1}{c|}{0.8819} & \multicolumn{1}{c|}{0.9750}   & \multicolumn{1}{c|}{0.63}        &  &  &  \\ \cline{1-7}
\multicolumn{1}{|c|}{\begin{tabular}[c]{@{}c@{}}$\lambda_{1}$/$\lambda_{2}$=1\\ Dynamic\end{tabular}}   & \multicolumn{1}{c|}{0.9950}    & \multicolumn{1}{c|}{0.9950} & \multicolumn{1}{c|}{0.9949} & \multicolumn{1}{c|}{0.9771} & \multicolumn{1}{c|}{0.9950}   & \multicolumn{1}{c|}{0.59}        &  &  &  \\ \cline{1-7}
\multicolumn{1}{|c|}{\begin{tabular}[c]{@{}c@{}}$\lambda_{1}$/$\lambda_{2}$=2\\ Dynamic\end{tabular}}   & \multicolumn{1}{c|}{1.0000}    & \multicolumn{1}{c|}{1.0000} & \multicolumn{1}{c|}{1.0000} & \multicolumn{1}{c|}{1.0000} & \multicolumn{1}{c|}{1.0000}   & \multicolumn{1}{c|}{0.56}        &  &  &  \\ \cline{1-7}
\multicolumn{1}{|c|}{\begin{tabular}[c]{@{}c@{}}$\lambda_{1}$/$\lambda_{2}$=0.5\\ Vote\end{tabular}}    & \multicolumn{1}{c|}{0.9757}    & \multicolumn{1}{c|}{0.9750} & \multicolumn{1}{c|}{0.9736} & \multicolumn{1}{c|}{0.8819} & \multicolumn{1}{c|}{0.9750}   & \multicolumn{1}{c|}{0.12}        &  &  &  \\ \cline{1-7}
\multicolumn{1}{|c|}{\begin{tabular}[c]{@{}c@{}}$\lambda_{1}$/$\lambda_{2}$=1\\ Vote\end{tabular}}      & \multicolumn{1}{c|}{0.9950}    & \multicolumn{1}{c|}{0.9950} & \multicolumn{1}{c|}{0.9949} & \multicolumn{1}{c|}{0.9771} & \multicolumn{1}{c|}{0.9950}   & \multicolumn{1}{c|}{0.08}        &  &  &  \\ \cline{1-7}
\multicolumn{1}{|c|}{\begin{tabular}[c]{@{}c@{}}$\lambda_{1}$/$\lambda_{2}$=2\\ Vote\end{tabular}}      & \multicolumn{1}{c|}{1.0000}    & \multicolumn{1}{c|}{1.0000} & \multicolumn{1}{c|}{1.0000} & \multicolumn{1}{c|}{1.0000} & \multicolumn{1}{c|}{1.0000}   & \multicolumn{1}{c|}{0.05}        &  &  &  \\ \cline{1-7}
\multicolumn{1}{l}{}                                                                & \multicolumn{1}{l}{}           & \multicolumn{1}{l}{}        & \multicolumn{1}{l}{}        & \multicolumn{1}{l}{}        & \multicolumn{1}{l}{}          & \multicolumn{1}{l}{}             &  &  &  \\
\multicolumn{1}{l}{}                                                                & \multicolumn{1}{l}{}           & \multicolumn{1}{l}{}        & \multicolumn{1}{l}{}        & \multicolumn{1}{l}{}        & \multicolumn{1}{l}{}          & \multicolumn{1}{l}{}             &  &  & 
\end{tabular}
\vspace{-0.4in}
\end{table}

\section{Conclusion and Future Directions}
\label{sec:conclusion}
We have proposed a federated learning (FL)-backed system for fake task detection in mobile crowdsensing (MCS). Two FL-based models are implemented with different ML algorithms and different local datasets in participating devices where the aggregated decision leverages risk/loss-awareness with utilities and losses with respect to task values. A traditional centralized platform with ensemble ML algorithms (e.g., Adaboost, Bagging, and XGBoost) for detection is introduced to compare with an FL-based system. In the centralized system, experimental results show Adaboost and XGBoost perform the highest G-mean $0.9743$ and accuracy $0.9900$. The proposed approach outperforms the prior work in \cite{chen2020deep} with an improvement of $12\%$ in G-mean and $7.3\%$ in overall accuracy. 
In the FL-based model with a large dataset, the best performance is the same as a centralized system with G-mean $0.9743$ and accuracy $0.9900$ in dynamic and vote based cases. On the other hand, under smaller datasets, the federated model ensures up to $100\%$ overall accuracy in both dynamic and static reputation cases with the loss incurred due to false prediction of legitimate and illegitimate tasks. Higher detection performance also leads to lower average loss in terms of task values for dynamic reputation and vote-based federated models. We are currently generalizing FL-based model to other datasets to verify performance with higher number of devices.



\bibliographystyle{plain}

\bibliography{refs.bib}

\end{document}